\documentclass{article}

\usepackage{amsmath}
\usepackage{amssymb}
\usepackage{graphicx}
\usepackage{geometry}
\title{Can machines think efficiently?}
\author{Adam Winchell}
\date{} 

\begin{document}

\maketitle

\begin{abstract}
The Turing Test is no longer adequate for distinguishing human and machine intelligence. With advanced artificial intelligence systems already passing the original Turing Test and contributing to serious ethical and environmental concerns, we urgently need to update the test. This work expands upon the original imitation game by accounting for an additional factor: the energy spent answering the questions. By adding the constraint of energy, the new test forces us to evaluate intelligence through the lens of efficiency, connecting the abstract problem of thinking to the concrete reality of finite resources. Further, this proposed new test ensures the evaluation of intelligence has a measurable, practical finish line that the original test lacks. This additional constraint compels society to weigh the time savings of using artificial intelligence against its total resource cost.
\end{abstract}

\section*{The Turing Test needs to be generalized}
Alan Turing asks "Can machines think?" in his seminal work$^1$-- given the proliferation of artificial intelligence and the rapid pace of innovation, there is an immediate need to update Turing's question to help delineate human versus machine in this modern day.

One may argue it is a wasted effort to move on to an "improved" Turing Test when we are still scratching the surface of the original. However, as we already have evidence of machines passing the Turing Test,$^{2,3,4}$ decision makers need additional tools to help navigate the ethical and philosophical quandaries that will continue to arise.

Others may argue the Turing Test is no longer relevant. There has been ample discussion that the Turing Test was never meant to be a practical test, and that focusing on it distracts us from the more serious issues of safety and reliability;$^5$ while others advocate to focus on efforts to create new sorts of tests that examine reasoning such as the Abstraction and Reasoning Corpus.$^6$

Although I see merit in both of these arguments, the key aspect missing from both is the approachability to the layman. The accessibility of the original Turing Test is unparalleled and its societal impact in guiding the conversation around artificial intelligence has been essential.

Humans are increasingly being misled and deceived by machines, requiring governments and scientists to grapple with the Turing Test's failures and their implications (e.g. large language models enabling suicidality and self-harm$^7$). The growing reliance on sophisticated intelligent systems leads to direct impact on energy costs for consumers, not to mention the strain on water resources, etc.$^8$ We are staring down the cliff of the enormous energy consumption of these intelligent systems and it is essential for us as a society to weigh more than just the time savings that these tools afford us.

\section*{A Motivating Example}
Suppose we wanted to create three versions of the internet: one for just humans, one for just machines, and another for humans and machines. In order to do so, we would need to verify the class of the participating entity.

In order to assess whether an entity is a machine, one might ask them to: "pick out the pixel from a large image presented as a matrix that has hex color F54927 in under 5 seconds"; a trivial task for a machine, but an impossible one for a human. This is an example that can be asked in the original formulation of the imitation game.

This question is phrased as a matter of ability: can the agent do the task, or not? However, this phrasing obscures a subtlety to the question. Without loss of generality or gain of specificity, we can rephrase it as: can the agent do the task efficiently enough to meet a time constraint? In other words, it is a question about the agent’s efficiency.

However, we can't always measure things in time (e.g. examining historical posts on social media), so what is more general? Given that we are discussing digital spaces, Landauer's principle$^9$ is relevant as it sets a theoretical lower bound of the energy consumption to perform the computation (e.g. interacting on the internet). This suggests that we need to somehow measure the energy involved for the agent to complete the task.

Consider differentiating humans and machines on the internet: as all digital communication is just a stream of $0$s and $1$s, we cannot differentiate a human versus a machine if they send out the same stream of $0$s and $1$s. Imagine further a human copying a machine's output. Without knowing the energy consumption to generate the output, this would be indistinguishable from an output produced by just a human.

These are examples of many-to-one mappings, e.g. two users can enter the same information that is posted to the internet. Therefore in order to differentiate the output, we need to know something about how the input was entered. If we had the ability to measure the energy involved to create the stream of $0$s and $1$s, then we could differentiate humans and machines.

\section*{The Energy Efficient Imitation Game}
I propose we consider a different question: "Can machines think efficiently?" I will, like Turing in his paper, note the arduousness of meaningfully and accurately defining the terms "machines" and "think", and consider a more tractable version of this question.

Consider an "energy efficient imitation game". It is played with three players: the liar, the truthteller, and the interrogator. The liar and truthteller are in a room together, while the interrogator is in a separate room, unable to directly interact with the other players.

The liar’s goal is to deceive the interrogator into believing that they are the truthteller. The liar can, of course, lie.

The objective for the truthteller is to convince the interrogator that they are in fact the truthteller. The truthteller cannot lie.

The interrogator’s goal is to determine which player is the liar by asking a set of questions. Questions are asked through an intermediary so no information can be gleaned other than the content of the answer and the associated amount of energy it took to answer the question.

The interrogator has access to a device that measures the energy expended by a player when answering each question. In order to simplify further discussion, let's call this device a \textit{psychoergometer}; for the literarily inclined, the word is a combination of the Greek words for mind (psyche), work (erg), and measure (meter). The psychoergometer is able to measure energy output in both humans and machines, and it is able to report those energy measurements in the same unit of measurement.

The interrogator knows the two players by labels X and Y, and at the end of the game says either "X is the liar" or "Y is the liar".\newline

An example question:\newline

Interrogator: Will X please tell me the length of their hair?\newline

Now suppose X is actually the liar, their answer might therefore be: \newline

Liar: I am bald.\newline

As Turing originally supposed, we now ask the question, "What will happen when a machine takes the part of the liar in this game?" Additionally, "How will the energy measurements influence the interrogator's judgements?" These more precise questions replace our original question, "Can machines think efficiently?"

While we very well could stop there and consider the ramifications of this extended problem statement, there are two more additional dimensions to the problem that are necessary to consider:

\begin{enumerate}
    \item What questions are we asking during the game?
    \item How do different sets of players change our perspective on the game?
\end{enumerate}

Why are these additional questions worth investigating? The answer, I argue, lies in the fundamental nature of specialization and trust.

\section*{The Imperative of Specialization and Trust}
The key to efficiency often lies in specialization.$^{10}$ A drone engineered for a single task, such as monitoring a solar farm, is likely to perform that one job more efficiently than a multi-purpose drone designed to conduct search-and-rescue, aerial photography, and surveillance. Within the biological world, we see the same trade-offs: crows are skilled at using simple tools and solving sequential problems. Should we assign a human to these tasks if a crow can do it with less energy?

This notion of specialization extends directly to the question of delineating intellectual entities. One may ask, why is it important to segregate entities based on their fundamental nature and specialized efficiency? The answer boils down to trust.

It is inherent for humans to split the world into "us vs. them",$^{11}$ as there are certain affordances one obtains when you know you are working with "your" group. Take for example the journal \textit{Nature}; because the journal has strict entrance criteria for acceptance, I can safely cite any work and build off of it (with the understanding that science is always evolving and mistakes are possible). Trust drastically reduces the speed of verification of data – if a machine learning model wants to improve on a task that is impossible for humans to perform, filtering out human related data is an essential first step.

Let's return to the earlier argument about creating three versions of the internet: one for just humans, one for just machines, and another for humans and machines. I argue that to create a human-only section of the internet requires playing the energy efficient imitation game; the game requires psychoergometers; psychoergometers do not exist; and if they did, it would, for obvious reasons, be impractical to use them at scale. Therefore I argue that creating a human-only version of the internet is not feasible. What I am about to supply is not a mathematical proof in any sense, but I do offer a sort of logic that underpins this belief.

Large language models blur the distinction between bot and human on the internet, so the imitation game, as opposed to the energy efficient imitation game, is insufficient for differentiability (Completely Automated Public Turing test to tell Computers and Humans Apart (CAPTCHA) is nearly fully broken by bots$^{12}$).

Suppose there exists some other condition (e.g. not energy efficiency) that we can add to the imitation game in order to differentiate humans and machines. Consciousness is a great example: suppose the interrogator has access to a device that allows them to measure the consciousness of a player (a \textit{psychoencephalometer} as it were). This same problem of "define a metric and measure it" begs the question, do we think one metric is inherently easier to scale to all humans and machines? My gut take is no, there is not one magic metric. Like everything in life, there are pros and cons to every choice, and those cons mean the coverage of the questions of the imitation game variant being played will never be $100\%$.

So if delineating the internet into human vs. machine requires playing the energy efficient imitation game, what can we change about the game to bring it back into the realm of feasibility?

Suppose instead we wanted to create three versions of the internet: one for just quantum computers, one for just machines (classical and quantum), and another for humans and machines. We already showed through the hypothetical of the pixelated image that we can assess whether an entity is a machine. To assess if an entity is a quantum computer, we could ask to factor a very large number into its prime constituents in under $5$ seconds; a trivial task for quantum computers due to Shor's algorithm$^{13}$ but an impossible one for classical computers and humans.

\section*{Critique of the New Problem}
Borrowing from Turing, "What is the answer to this new form of the question," one may also ask, "Is this new question a worthy one to investigate?" Why should we care about the energy efficient imitation game?

This may feel trite, but one simple answer to the worthiness question is by asking another question "Should we use the right tool for the job?" Putting it another way, I am postulating: humans want to answer as many questions as possible with as few resources as possible.$^{14}$ This framing of the problem distinguishes between the quality of an answer and the cost of obtaining it. Consider the following situation to help motivate this distinction:\newline

Question: Add $34,957$ to $70,764$.\newline

Calculator Answer: (Near immediate) $105,721$.\newline

Human Answer: (Pauses about 30 seconds and then gives as answer) $105,721$.\newline

Large Language Model Answer: (Routes question to math engine and then answers) $105,721$.\newline

As Turing originally noted, the question and answer method seems to be suitable for introducing almost all fields of human endeavor. We do not want to penalize machines for their poor performance in beauty competitions, nor to penalize a human for losing in a race against an airplane. We, as Turing, pick the question-answer format because it makes these irrelevant differences irrelevant. This allows us to exclude a long discussion on the field of robotics and its interplay with that of artificial intelligence. Crucially, by eliminating irrelevant physical attributes, the question-answer method allows us to assess artificial intelligence's performance in its primary domain: digital spaces.

It might be argued that playing the energy efficient Turing game is already trivial in many circumstances even without the use of a psychoergometer: a human cannot in a reasonable amount of time classify pixels in an image compared to a machine as already discussed. However, one could easily imagine a plethora of situations in which the decision over which tool to leverage (human, machine, quantum intelligence, or other) is much less obvious. One of the goals of this work is to help identify the contours of this as yet nebulous problem space and identify places in which discussion might fruitfully begin – the goal is to help uncover the exceptions that prove the rule.

We can actually embed the original imitation game within this new game by asking "Can machines think?" A player could pontificate on the definition of intelligence, the scientific and cultural touchstones that bias this question, or answer in any one of an infinite number of ways. The distribution of energy being spent answering the question across the population would be as informative as (if not more so than) the answers themselves. I would wager that the large majority of humans would give a short simple answer, whereas current machines would provide large expositions reflecting that of a philosopher. For questions that don't have definitive answers, the journey to the answer is far more interesting than the answer itself.

\section*{Contrary Views on the Main Question}
With the game defined and the groundwork of the arguments laid out, we can now debate whether or not machines can in fact think efficiently, the variant, and the related questions from the previous sections.

I will start with my beliefs on the matter, followed by considering opinions opposed to my own.

As an avid reader of science fiction and follower of current scientific achievements, I often like to imagine what the future of human society, earth, and more broadly the universe will look like in the future. I have no doubt in my mind that at some point in the future, humanity will have helped create artificial general intelligence (AGI), and that AGI will eventually be as energy efficient as (if not more so than) humans on most tasks. However, until machines are as efficient as humans, then the energy efficient imitation game is an essential exercise for us to grapple with as we navigate our lives trying to answer questions. Why should I use a non-deterministic expensive machine (large language model) to do arithmetic when there is a deterministic cheap tool (calculator) able to better do the job?

Although I'm nervous about the future of humanity, my optimism allows me to envision that humans and machines will depend on each other in ways we have yet to discover. If for some example, there exists scientific phenomena in deep space that would be prohibitively difficult for inorganic beings (my imagined AGI machine) to explore, but would be much easier for organic beings, then the future of humans and machines may be bright after all.

I have yet to mention that the form and function of energy being consumed is also a key factor in considering the right entity for the job. As of the time of writing this paper, battery technology in cars has not yet overcome the trials and tribulations of winter; so while you may think electric vehicles may be more efficient in getting one around in general, I live in a place with cold winters and no garage, and I will stick to gasoline cars as they are more efficient for my needs and circumstances.

Another critical consideration that often gets overlooked is the efficiency of information storage and retrieval. The energy cost of providing an answer does not start at the moment the entity begins to think; it includes the energy cost of maintaining the knowledge required for the answer. For instance, a system that uses a $4$-bit system to store information has a higher energy cost than a $2$-bit system; if our task is to observe the outcome of flipping a single coin and reporting the outcome in $1$ year, we should use the $2$-bit system for this task as it costs less energy to obtain the desired result.

Perhaps our thinking on this topic is too shallow. Maybe in the far flung future, humans, classical computer AGI, and quantum computer AGI will work in tandem, each performing the role for which it is most efficient.

What the energy efficient imitation game does, in contrast to the original Turing Test, is introduce an ultimate constraint on the interrogator. By including the cost of energy in the exchange, it ensures there are only a finite number of questions the human can ask before their energy runs out (e.g. the human dies), which is not the case in the original formulation. This guarantees an end to the game, providing a measurable and practical finish line that the original test lacked.

As Turing once did, let us now proceed to consider opinions opposed to my own.

\subsection*{Objection: Machines will never be smarter than humans}
As Turing noted this in the 'The "Heads in the Sand" Objection', I am not here to convince anyone of the limits of machines' abilities, nor am I here to prop up humans on a pedestal and say they can never be fully replaced. I am speaking to the people who are concerned about our shared future with intelligent machines. I am speaking to the people who like to engage in intellectual debates in good faith.

\subsection*{Objection: A mathematical argument is better}
An energy-based assessment of intelligence is nothing new.$^{15,16,17}$ So, at first glance, embarking on a persuasive essay format seems fruitless when a rigorous alternative is already present. As someone who has training in mathematics, I almost didn't write this piece as the mathematics felt like an open and closed case so-to-speak.

But then I started talking to my friends (who hail from many diverse backgrounds), and was startled to learn many of them (without any background in mathematics, computer science, etc.) had read Turing's paper and, more to the point, felt like they were able to grapple with its core thesis in terms that felt relatable to them.

There is always a tension between accuracy and narrative within scientific communication, and while I am not trying to dissuade a more rigorous approach to this topic, it is essential that the public, the policy makers, and those raising the next generation be given the opportunity to peek behind the curtain and participate in discourses that actively shape all of our lives.

\subsection*{Objection: The moral hazard of dehumanization}
By incorporating efficiency into the estimate of an entity's utility, it seems at first glance that I am promoting a utilitarian, dehumanizing view. If a human is less efficient than a machine at a certain task, my logic seems to imply the human should be replaced.

The purpose of this new test is to help identify the right tool for the job, but right is a loaded word. We already live in an era where people seek therapy from machines$^{18}$ which may very well be the most energy efficient choice, however the value of humanness cannot be overlooked. It is beyond the scope of this essay to try to quantify humanness,$^{19}$ but I think it is sufficient to say there will always exist a desire for human to human interaction that cannot be replaced regardless of the energy costs associated.

\section*{Learning Machines Revisited}
Unlike Turing, we do not need to dwell on the feasibility and ability to create such learning machines – we live in a society implicitly and explicitly driven by such machines. As closing thoughts, let us return to psychoergometers and the idea of a hierarchy of intelligence.

Although I already said that a psychoergometer is impossible in practice, we could reduce our requirements and measure only part of the energy involved in answering a task. If we use a pseudo-psychoergometer, a pared-down version of the device that measures a fraction of the total energy spent—then we might be able to play an approximate energy-efficient imitation game similar to CAPTCHA. A future essay should be dedicated to pseudo-psychoergometers as we need practical solutions to these problems.

I would like to offer the reader an opportunity to think on the idea of a hierarchy of intelligence. As described in this essay, if we carefully craft our set of questions and play the energy efficient imitation game we can determine a hierarchy of intelligence for that set of questions.

Is there a set of questions where humans are always at the top of the hierarchy?

Is the above question unanswerable due to Gödel's incompleteness theorem?$^{20}$

To answer these questions mathematically requires us to be precise and write down a definition for "human".

There are two sensible options: humans are Turing Machines$^{21}$ or humans are Hypercomputational machines (such as those described by John Lucas$^{22}$ and Roger Penrose$^{23}$).

If we assume humans are Turing Machines, there is an additional question we must ask. Is the set of
questions that will be used in the energy efficient imitation game an infinite set?

If the set of questions is infinite (i.e. you are assuming that humans can outlive the biological constraints$^{24}$) and each participant has unbounded computation,
then the answer to whether a strict hierarchy exists is undecidable in general. Determining whether one Turing machine definitively beats another across every question is undecidable in general. The Halting Problem can be reduced to this comparison problem.

If the set of questions is finite, thus aligning with human’s biological constraints, and each participant is given a finite execution budget, then the comparison becomes decidable. Even a Turing machine with an unbounded tape can access only finitely many memory cells within that budget. We can therefore run each participant until it answers or exhausts its permitted resources, record its performance, and construct the resulting partial hierarchy. Under these constraints, the hierarchy for this finite game is decidable.

If we assume instead that humans are hypercomputational machines, then, by definition, there exists a set of questions that humans can answer but that no Turing machine can answer in full.

Therefore the decidability of the original question, \textit{Is there a set of questions where humans are always at the top of the hierarchy?}, depends on your choice of computational model and resource constraints.

\section*{Acknowledgements}
I would like to thank: Hunter Wapman, Julia Versel, Micah Ross, Tyler Scott, Jessica Finocchiaro, Gabe O'Brien, Pierre Delcourt, Behzad Hasani, Avinash Saraf.

\section*{Competing interests}
The author declares the following competing interest:\newline

\noindent The author works at Google. The views and research presented in this paper are the author's own and do not reflect the official policy or position of the author's employer.

\section*{Additional Information}
Correspondence and requests for materials should be addressed to \texttt{adamwinchellresearch@gmail.com}.


\begin{thebibliography}{99} 
\bibitem{Turing50} Turing, A. M. Computing machinery and intelligence (1950). \textit{Mind}, 59(236), 33-60.
\bibitem{Jones25} Jones, Cameron R., and Benjamin K. Bergen. "Large language models pass the turing test." \textit{arXiv preprint} arXiv:2503.23674 (2025).
\bibitem{Biever23} Biever, C. (2023). ChatGPT broke the Turing test-the race is on for new ways to assess AI. \textit{Nature}, 619(7971), 686-689.
\bibitem{Mei24} Mei, Q., Xie, Y., Yuan, W., \& Jackson, M. O. (2024). A Turing test of whether AI chatbots are behaviorally similar to humans. \textit{Proceedings of the National Academy of Sciences}, 121(9), e2313925121.
\bibitem{Gibney} Gibney, E. AI language models killed the Turing test: do we even need a replacement?. \textit{Nature}.
\bibitem{Chollet19} Chollet, F. (2019). On the measure of intelligence. \textit{arXiv preprint} arXiv:1911.01547.
\bibitem{Schoene25} Schoene, A. M., \& Canca, C. (2025). For Argument's Sake, Show Me How to Harm Myself!': Jailbreaking LLMs in Suicide and Self-Harm Contexts. \textit{arXiv preprint} arXiv:2507.02990.
\bibitem{Siddik21} Siddik, M. A. B., Shehabi, A., \& Marston, L. (2021). The environmental footprint of data centers in the United States. \textit{Environmental Research Letters}, 16(6), 064017.
\bibitem{Landauer61} Landauer, R. (1961). Irreversibility and heat generation in the computing process. \textit{IBM journal of research and development}, 5(3), 183-191.
\bibitem{Hunt95} Hunt, S. D., \& Morgan, R. M. (1995). The comparative advantage theory of competition. \textit{Journal of marketing}, 59(2), 1-15.
\bibitem{Cikara11} Cikara, M., Botvinick, M. M., \& Fiske, S. T. (2011). Us versus them: Social identity shapes neural responses to intergroup competition and harm. \textit{Psychological science}, 22(3), 306-313.
\bibitem{Dinh23} Dinh, N. T., \& Hoang, V. T. (2023). Recent advances of Captcha security analysis: a short literature review. \textit{Procedia Computer Science}, 218, 2550-2562.
\bibitem{Shor94} Shor, P. W. (1994, November). Algorithms for quantum computation: discrete logarithms and factoring. In \textit{Proceedings 35th annual symposium on foundations of computer science} (pp. 124-134). Ieee.
\bibitem{Rothe18} Rothe, A., Lake, B. M., \& Gureckis, T. M. (2018). Do people ask good questions?. \textit{Computational Brain \& Behavior}, 1(1), 69-89.
\bibitem{Perrier25} Perrier, E. (2025, August). Watts-Per-Intelligence: Part I (Energy Efficiency). In \textit{International Conference on Artificial General Intelligence} (pp. 46-57). Cham: Springer Nature Switzerland.
\bibitem{Ganesh18} Ganesh, N. (2018, November). Thermodynamic intelligence, a heretical theory. In \textit{2018 IEEE International Conference on Rebooting Computing (ICRC)} (pp. 1-10). IEEE.
\bibitem{Mehdi24} Mehdi, S., \& Tiwary, P. (2024). Thermodynamics-inspired explanations of artificial intelligence. \textit{Nature Communications}, 15(1), 7859.
\bibitem{Kuhail25} Kuhail, M. A., Alturki, N., Thomas, J., Alkhalifa, A. K., \& Alshardan, A. (2025). Human-human vs human-AI therapy: An empirical study. \textit{International Journal of Human–Computer Interaction}, 41(11), 6841-6852.
\bibitem{Haslam05} Haslam, N., Bain, P., Douge, L., Lee, M., \& Bastian, B. (2005). More human than you: attributing humanness to self and others. \textit{Journal of personality and social psychology}, 89(6), 937.
\bibitem{Godel92} Gödel, K. (1992). On formally undecidable propositions of Principia Mathematica and related systems. Courier Corporation.
\bibitem{Turing36} Turing, A. M. (1936). On computable numbers, with an application to the Entscheidungsproblem. J. of Math, 58(345-363), 5.
\bibitem{Lucas61} Lucas, J. R. (1961). Minds, machines and gödel1. Philosophy, 36(137), 112-127.
\bibitem{Penrose16} Penrose, R. (2016). The emperor's new mind: Concerning computers, minds, and the laws of physics. Oxford University Press.
\bibitem{Sandberg08} Sandberg, A., Bostrom, N. (2008). Whole brain emulation: A roadmap.


\end{thebibliography}
\end{document}